# Defeats GAN: A Simpler Model Outperforms in Knowledge Representation Learning

Heng Wang
School of Data and Computer Science
Sun Yat-sen University
Guangzhou, China
e-mail: wangh376@mail2.sysu.edu.cn

Mingzhi Mao
School of Data and Computer Science
Sun Yat-sen University
Guangzhou, China
e-mail: mcsmmz@mail.sysu.edu.cn

*Abstract*—The goal of knowledge representation learning is to embed entities and relations into a low-dimensional, continuous vector space. How to push a model to its limit and obtain better results is of great significance in knowledge graph's applications. We propose a simple and elegant method, Trans-DLR, whose main idea is dynamic learning rate control during training. Our method achieves remarkable improvement, compared with recent GAN-based method. Moreover, we introduce a new negative sampling trick which corrupts not only entities, but also relations, in different probabilities. We also develop an efficient way, which fully utilizes multiprocessing and parallel computing, to speed up evaluation of the model in link prediction tasks. Experiments show that our method is effective.

*Knowledge representation; dynamic learning rate; negative sampling; multiprocessing; parallel computing*

## I. INTRODUCTION

Knowledge Graph is a directed graph structure which is composed of various kinds of entities and their relations in our world. Typical knowledge graphs include Wordnet [1], Freebase [2], Yago [3], to name a few. Knowledge graph is playing a pivotal role in many NLP applications, such as relation extraction [4], question answering [5], and social network mining [6].

Facts in a knowledge graph are commonly represented as triples (*head entity, relation, tail entity*), abbreviated as (*h, r, t*). They are obtained by human labor, rules or distant supervision [7], which are usually far from complete. Knowledge graph representation aims to represent entities and relations as symbols, numbers, or vectors, aiding in completing missing links and finding new facts for a knowledge graph. Inspired by [8], a great deal of effort have been made to embed entities and relations into a low-dimensional, continuous vector space, such as [9-12], with different loss functions adopted. Let $\Delta$ denote all the triples in a knowledge base. A triple $(h, r, t)$ is positive if $(h, r, t) \in \Delta$, otherwise negative if $(h, r, t) \notin \Delta$. The basic ideas behind these models is that the loss of negative triples should be at least $\gamma$ greater than the loss of positive triples, which is known as margin loss. Readers can refer to Section II to get more detailed introduction.

During training, all the models mentioned above suffer from the problem of local optimum and inability to step forward in performance (See Fig. 1). How to push a model to its limit and learn a better representation is of great significance in knowledge graph's downstream applications. Recently, [13] proposes a knowledge embedding framework which utilizes GAN in negative sampling, called **Trans-GAN**, to mine the potential of models by generating high-level negative samples. However, it has several drawbacks. Firstly, GAN often faces the problem of non-convergence or collapse in training, leading to a poor result when it happens. Secondly, GAN consists of generator and discriminator networks, which needs more parameters.

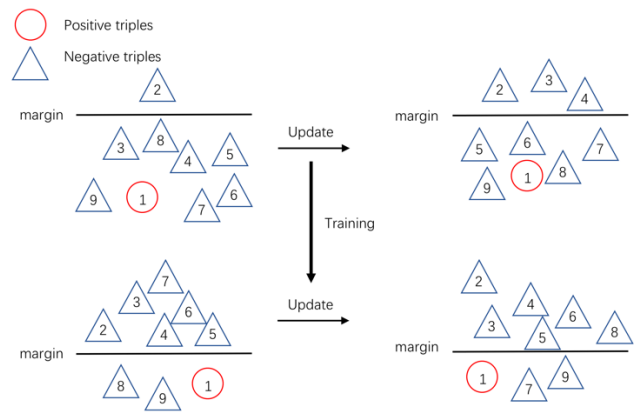

Figure 1. Illustration of local optimum in training. After training for a while, the model pushes nearly the same number of negative triples out of and into the margin, resulting in no improvement of performance.

In this paper, we propose a simpler and more elegant method whose main idea behind is dynamic learning rate (**DLR**). Experiments show that our DLR-based methods outperforms GAN-based methods remarkably under most circumstances.

Our contributions in this paper are as follows:
- We incorporate DLR in knowledge representation learning which can dynamically adjust the learning rate of a model, pushing the model to a better optimum.
- We propose a new negative sampling method which not only corrupts entities, but also relations in different probabilities. So the model can learn better representation for both entities and relations.

- We design a method, which fully utilizes multiprocessing and parallel computing, to speed up the evaluation of a model in link prediction tasks. It saves a lot of time and makes DLR algorithm more applicable.

## II. RELATED WORK

We define several mathematical notations before proceeding. Denote a triple by $(h, r, t)$ and their corresponding vectors by $\{\mathbf{h}, \mathbf{r}, \mathbf{t}\}$; matrices by bold upper case letters like $\mathbf{M}$. $f(\mathbf{h}, \mathbf{r}, \mathbf{t})$ refers to loss function, for which positive triples scores lower and negative triples scores higher. Models for knowledge representation learning can be roughly divided into two groups: translation-based and non-translation-based.

### A. Translation-Based Models

**TransE.** TransE [9] treats $r$ as translation from $h$ to $t$ in the same vector space. The loss function is
$$f(\mathbf{h}, \mathbf{r}, \mathbf{t}) = \|\{\mathbf{h} + \mathbf{r} - \mathbf{t}\}\|_{l_1/l_2}. \quad (1)$$
**TransH.** To better handle 1-to-N, N-to-1, and N-to-N relations, [10] proposes TransH, which views $r$ as translation on a hyperplane. An entity has different projections according to different relations. Denote the norm vector of $r$ by $\mathbf{w_r}$, entities are projected as
$$\mathbf{h}_\perp = \mathbf{h} - \mathbf{w_r}^\top \mathbf{h} \mathbf{w_r}, \quad \mathbf{t}_\perp = \mathbf{t} - \mathbf{w_r}^\top \mathbf{t} \mathbf{w_r}, \quad (2)$$
and the loss function is defined as
$$f(\mathbf{h}, \mathbf{r}, \mathbf{t}) = \|\mathbf{h}_\perp + \mathbf{r} - \mathbf{t}_\perp\|_{l_1/l_2}. \quad (3)$$
**TransR.** In TransE and TransH, entities and relations share the same vector space, so $\mathbf{h}$, $\mathbf{r}$ and $\mathbf{t}$ require the same size of dimension. TransR [11] allows entity space and relation space to have different sizes, and use $\mathbf{M_r}$ as the projection matrix from entity space to relation space. Now the loss function becomes
$$f(\mathbf{h}, \mathbf{r}, \mathbf{t}) = \|\mathbf{M_r}\mathbf{h} + \mathbf{r} - \mathbf{M_r}\mathbf{t}\|_{l_1/l_2}. \quad (4)$$
**TransD.** TransD [14] extends TransR by constructing projection matrices according to not only relations, but also entities. It is defined as
$$\mathbf{M_{rh}} = \mathbf{r_p}\mathbf{h_p}^\top + \mathbf{I}, \quad \mathbf{M_{rt}} = \mathbf{r_p}\mathbf{t_p}^\top + \mathbf{I}, \quad (5)$$
where $\mathbf{p}$ marks projection vectors. Now the loss function turns into
$$f(\mathbf{h}, \mathbf{r}, \mathbf{t}) = \|\mathbf{M_{rh}}\mathbf{h} + \mathbf{r} - \mathbf{M_{rt}}\mathbf{t}\|_{l_1/l_2}. \quad (6)$$
Other translation-based models include TransG [15], TransA [16], etc, with definition of spaces, ways of projection and loss functions different.

### B. Non-Translation-Based Models

**Unstructured Model** [17-18] is a simple case of TransE which always views $\mathbf{r}$ as $\mathbf{0}$. **Structured Embedding (SE)** [19] assigns two projection matrices $\mathbf{M_{r,h}}$ and $\mathbf{M_{r,t}}$ to represent a relation $\mathbf{r}$, with translation vectors $\mathbf{r}$ still $\mathbf{0}$. **Single Layer Model (SLM)** [20] fine-tunes SE by adding relation vector $\mathbf{u_r}$ for each relation and nonlinear transformation. **Neural Tensor Network (NTN)** [20] is more expressive than SLM since it considers second-order correlations into nonlinear transformation.

Thanks to the prosperity of deep learning, some models based on neural networks have also been developed in recent years, such as ProjE [12] and ConvE[21].

## III. OUR METHOD

To handle the problem of local optimum in which training stuck, we propose two kinds of DLR algorithms which can be incorporated into random initialization and pretrained initialization models, respectively. We also develop a new way to generate negative samples which replaces entities and relations in different probabilities.

### A. Trans-DLR/Trans-DLR2

The basic idea behind DLR is that when a model converges to local optimum, it cannot distinguish new negative triples at all, leaving only pushing nearly equal number of distinguished negative triples out of and into margin. Analogize with climbing hills, a model stuck in local optimum is like a man jumping along the same contour, unable to step higher. However, if we lower the learning rate after the performance has no breakthrough for some epochs, the model can be fine adjusted and learn new features of triples, so new negative triples may be distinguished and pushed out of the margin. This is like a man taking smaller step and jumping to a higher attitude. We simply name this algorithm **Trans-DLR**, with detailed process shown in Algorithm 1.

---
**Algorithm 1** Trans-DLR
---
**Input:** Number of epochs per evaluation on validation set $E$; Number of times for decrease of learning rate (lr) $D$; Max times allowed for performance not improving $P$; Initial lr $\lambda$; lr decrease speed $s$; Training set $T = \{(h, r, t)\}$; Validation set $V = \{(h, r, t)\}$; Model $M$.
**Output:** Knowledge representation learned by $M$.
1:   Initialize $M$ and lr with $\lambda$;
2:   **loop**
3:      Train $M$ on $T$ and update;
4:      **for** every $E$ epochs **do**
5:         Evaluate model performance on a sample of $V$ and its negative samples;
6:         **if** No improvement in performance for consecutive $P$ times **then**
7:            **if** Times of lr decrease exceeds $D$ **then goto** 12;
8:            **else** Decrease lr by $s$;
9:         **end if**
10:     **end if**
11:     **end for**
12:  **end loop**

---

Algorithm 1 is suitable for a model whose parameters is initialized randomly, meaning the model is trained from scratch. However, for TransR and TransD which has a large size of parameters and prone to overfitting, initializing embeddings with results of TransE and projection matrices with identity matrices is needed. For the first dozens of epochs, the performance drops significantly, after which the model fine-tunes itself to get better and better results. If we simply apply Trans-DLR, the first epoch will be wrongly regarded as The best and the learning rate will continue dropping, even if the model hasn't adjust to its best under a certain learning rate.

So we improve Trans-DLR to **Trans-DLR2**, indicating the algorithm has two phases: dropping phase and tuning phase.

TABLE I. STATISTICS OF DATASETS USED IN THE EXPERIMENTS

| Dataset | #Rel | #Ent | #Train | #Valid | #Test |
|---|---|---|---|---|---|
| FB15k | 1,345 | 14,951 | 483,142 | 50,000 | 59,071 |
| WN18 | 18 | 40,943 | 141,442 | 5,000 | 5,000 |

Dropping phase lasts until no degrade in performance, and the final result of dropping phase is set as the baseline of tuning phase. Tuning phase is just like Trans-DLR. The learning rate of two phases can be initialized differently, with $\lambda_1$ fixed and $\lambda_2$ adjustable in training.

### B. Negative Sampling

Since negative samples don't exist in datasets, we need to generate them by ourselves. Previous methods only corrupt head or tail entities with equal probabilities or according to Bernoulli distribution (**Bern**) [10]. However, vectors involving relations should also be trained thoroughly, especially for TransH, TransR and TransD, which have more parameters of relations than entities. Therefore, we introduce a method which corrupts entities or relations in different probabilities. Denote the number of types of entities and relations by $\#Ent$ and $\#Rel$, respectively, and define $\#Total = 2 * \#Ent + \#Rel$. For each triple, we randomly replace head entity with probability $\#Ent/\#Total$, or tail entity $\#Ent/\#Total$, or relation $\#Rel/\#Total$. We abbreviate our method to **NSER** and previous method, which only corrupts entities, to **NSE**.

## IV. EXPERIMENTS AND ANALYSIS

We conduct experiments on two datasets proposed by [9]: FB15k and WN18, which are subsets of Freebase [2] and Wordnet [1], respectively. The former contains common facts of the world, and the latter is about lexical relations between words. Their statistics are listed in Table I, and #1-1, #1-N, #N-1, #N-N refers to the number of triples which contains one-to-one, one-to-many, many-to-one, many-to-many relations, respectively.

An important goal of learning knowledge graph embeddings is to predict the missing entity in a given triple, namely to predict $h$ given $(r,t)$ or $t$ given $(h,r)$. Generally two measurements are taken: $hits@10$ and $MeanRank$. $hits@10$ computes the proportion of ground truth entities whose scores are in top 10 among all entities, while $MeanRank$ averages the rank of ground truth entities for all test cases. Since some generated negative triples may be in the dataset, they should not be recognized as incorrect and need to be filtered out in training, validating and testing, so the final result will not be underestimated. We utilize $MeanRank$ as the measurement in validation.

Since Trans-DLR evaluates the model on the validation set quite frequently, how to speed up evaluation and save time is an important consideration. Conventionally, the evaluation algorithm scans triples one by one, predicts its head and tail entities, and calculates the rank. It has obvious drawbacks. On the one hand, it doesn't take advantage of matrix computation, which is often faster than looping on rows and columns. On the other hand, it doesn't fully utilize the computational resource on a computer with multi-core CPU. So, we design the evaluation method as follows. We only discuss our method on the problem of predicting the tail given head and relation, and the inverse problem, which predicts the head given tail and relation, is similar.

Let $c$ denote the number of CPU cores in a computer, $m$ denote the dimension of relation space, $n$ denote the number of triples in a subset, and $N$ denote the number of different entities in the whole dataset. For TransE, which has no relation-specific vector space, we first equally divide the evaluation / test set into $c$ parts. Then, we start $c$ processes, and each process deals with each part. In the process, it vertically stacks embeddings of head entities, tail entities, and relations, respectively, generating three $n * m$ matrices. We denote them by **H**, **T** and **R**. Add **H** and **R** to get **T**′, and calculate the pairwise distance between **T**′ and the set of all entities, getting an $n * N$ matrix. Now, we can find the ground truth entities' rank, and whether it's in top 10. Merge the results from $c$ parts, we can get $MeanRank$ and $hits@10$ of the whole evaluation / test set.

As for TransH, TransR, and TransD, which has relation-specific vector spaces, we first divide the evaluation / test set according to the relation, so each subset only contains triples with the same relation. Then, a list with $c$ processes is created, and all the subsets is arranged like a queue. Free process picks up the foremost subset in the queue. In the process, it first projects head / tail entities in the subset, and all the entities in the whole dataset to the vector space occupied by this relation. Under this new vector space, calculating $MeanRank$ and $hits@10$ is in a similar manner like what is mentioned above.

We conduct experiments on a 32-core CPU machine, with a GeForce GTX 1080 Ti Graphic Card (11G memory). The search space of parameters of Trans-DLR(2) are as follows: dimension $d \in \{20, 50, 100\}$, margin $\gamma \in \{1, 2, 4\}$, batch size $B \in \{30, 120, 480, 1440, 4800\}$, $s \in \{0.1, 0.3, 0.5, 0.7, 0.9\}$, $E \in \{1, 5, 10, 20\}$, $P \in \{1, 5, 10\}$, $D \in \{1, 5, 10, 20\}$, $\lambda, \lambda_1, \lambda_2 \in \{0.005, 0.001, 0.0005, 0.0001\}$.

After cross-validation, we set the hyperparameters as following. For FB15k, dimension $d = 100$, margin $\gamma = 1$, batch size $B = 4800$, $s = 0.5$, $E = 5$, $P = 5$, $D = 20$, $\lambda = 0.001$ using TransE and TransH, $\lambda_1 = 0.001$ and $\lambda_2 = 0.0005$ using TransR, $\lambda_1 = 0.001$ and $\lambda_2 = 0.001$ using TransD. Take $l_1$ as dissimilarity. For WN18, $d = 50$, $\gamma = 4$, $B = 1440$, $s = 0.5$, $E = 5$, $P = 5$, $D = 20$, $\lambda = 0.001$ using TransE and TransH, $\lambda_1 = 0.001$ and $\lambda_2 = 0.001$ using TransR, $\lambda_1 = 0.001$ and $\lambda_2 = 0.001$ using TransD. Take $l_1$ as dissimilarity. We choose Adam [22] as optimizer. Typically, the process terminates in 600 ~ 800 epochs.

[13] provides two types of GAN-based training methods, GAN-scratch and GAN-pretrain. The former trains the

discriminator and generator from scratch, while the latter uses generator to fine-tune the pretrained discriminator. Since our model are all trained from scratch, we compare our result with that of GAN-scratch. Evaluation results on link prediction of the whole test set are shown in Table II.

We discover that:

TABLE II. EVALUATION RESULTS ON LINK PREDICTION, COMPARED WITH TRANS-GAN.

| FB15k | Original | | Trans-GAN | | Trans-DLR(2) + NSE | | Trans-DLR(2) + NSER | | Trans-DLR(2) + NSE + Bern | | Trans-DLR(2) + NSER + Bern | |
|---|---|---|---|---|---|---|---|---|---|---|---|---|
| | Mean Rank | Hits@ 10(%) | Mean Rank | Hits@ 10(%) | Mean Rank | Hits@ 10(%) | Mean Rank | Hits@ 10(%) | Mean Rank | Hits@ 10(%) | Mean Rank | Hits@ 10(%) |
| TransE | 125 | 47.1 | 90 | 73.1 | 54 | 76.7 | 47 | 80.5 | 72 | 76.7 | 53 | 80.2 |
| TransH | 87 | 64.4 | 90 | 73.3 | 49 | 77.3 | 45 | 79.7 | 64 | 78.8 | 53 | 79.1 |
| TransR | 77 | 68.7 | 138 | 58.3 | 44 | 79.4 | 44 | 80.9 | 55 | 78.4 | 47 | 80.2 |
| TransD | 91 | 77.3 | 89 | 74.0 | 43 | 81.1 | 43 | 82.3 | 51 | 81.3 | 47 | 82.1 |
| WN18 | Original | | Trans-GAN | | Trans-DLR(2) + NSE | | Trans-DLR(2) + NSER | | Trans-DLR(2) + NSE + Bern | | Trans-DLR(2) + NSER + Bern | |
| | Mean Rank | Hits@ 10(%) | Mean Rank | Hits@ 10(%) | Mean Rank | Hits@ 10(%) | Mean Rank | Hits@ 10(%) | Mean Rank | Hits@ 10(%) | Mean Rank | Hits@ 10(%) |
| TransE | 251 | 89.2 | **244** | 92.7 | 307 | 93.3 | 285 | **94.1** | 245 | 93.0 | 281 | 93.1 |
| TransH | 303 | 86.7 | 276 | 86.9 | 279 | 93.5 | **262** | **94.1** | 265 | 93.0 | 295 | 93.6 |
| TransR | 225 | 92.0 | **213** | 87.3 | **213** | 93.3 | 250 | **93.8** | 225 | 92.8 | 240 | 93.1 |
| TransD | 229 | 92.5 | **221** | 93.0 | **221** | 94.0 | 234 | **94.1** | 245 | 93.1 | 262 | 93.7 |

1) For FB15k, Trans-DLR(2) significantly outperforms original and Trans-GAN in both MeanRank and hits@10. For WN18, Trans-DLR(2) outperforms Trans-GAN in hits@10, under all the four translation-based methods, while slightly underperforms in TransE, outperforms in TransH, and get a tie in TransR and TransD, in MeanRank.

2) Compared with NSE, there is stable improvement in performance with the usage of NSER, no matter whether Bern

trick is adopted. The improvement is more significant in FB15k since it contains much more relations and less entities.

3) The $MeanRank$ of Trans-DLR(2) under WN18 dataset doesn't obtain significant improvement. The reason might be that $\#Ent$ of WN18 is large while $\#Rel$ is small, i.e., it's a dense dataset. It is more difficult to embed a large number of entities into a limited-dimension vector space while keeping all the constraints satisfied. Moreover, during the later stage of training, the model may push an entity from top hundreds to top 10, with the cost of dropping another entity from top hundreds out of top thousands. Although $hits@10$ gets slight improvement, $MeanRank$ is unable to increase, which is like a dynamic equilibrium process.

4) Bern trick doesn't always assist the model when Trans-DLR(2) is adopted. The intention of Bern trick is to reduce false negative samples. However, this intended biased setting may render some entities not trained thoroughly. With Trans-DLR(2), the model is capable of learning subtle differences between positive and negative triples and reducing the interfere of false negative samples, while overcoming the flaw of Bern trick.

Table III shows the evaluation time per triple with and without using our method to speed up. After utilizing multiprocessing and parallel computing, evaluation on WN18 accelerates by 212x to 587x, and evaluation on FB15k even gains a speed up of at least 1,000x! This shows the effectiveness of our speed up method. Since FB15k contains much more relations than WN18, the CPU utilization rate

TABLE III. EVALUATION TIME PER TRIPLE WITH AND WITHOUT SPEEDING UP

| Dataset | Algorithm | Evaluation Time (ms) | |
|---|---|---|---|
| | | Speeding up | No Speeding up |
| FB15k | TransE | 0.54 | 890 |
| | TransH | 0.61 | 898 |
| | TransR | 1.22 | 1285 |
| | TransD | 0.60 | 851 |
| WN18 | TransE | 1.27 | 746 |
| | TransH | 3.79 | 804 |
| | TransR | 2.82 | 1064 |
| | TransD | 3.03 | 789 |

will be higher if one process only deals with one relation, so the acceleration rate is higher.

Table IV shows the amortized training time with using only the original model, adding Trans-DLR(2) and adding Trans-GAN. It can be seen that Trans-DLR(2) costs no more than twice time needed by the original model, while Trans-GAN requires at most 4x extra time, due to the complexity of GAN model. It can be inferred that our model is more time-efficient than Trans-GAN.

V. CONCLUSION AND FUTURE WORKS

In this paper, we introduce Trans-DLR and Trans-DLR2 to knowledge representation learning, which focuses on dynamic learning rate, applicable for random initialization and pretrained initialization, respectively. To make our DLR algorithm more feasible in applications, we also design an evaluation method based on multiprocessing and parallel computing, which fully utilizes computational resources and speeds up training and evaluation process a lot. Experiments show that our simple and elegant method outperforms

Trans-GAN under almost all circumstances. *We also propose NSER trick which is proved useful in this task.*

TABLE IV.  AMORTIZED TRAINING TIME PER EPOCH WITH DIFFERENT TRAINING WAYS

| Dataset | Algorithm | Evaluation Time (s) | | |
|---|---|---|---|---|
| | | Original | Trans-DLR(2) | Trans-GAN |
| FB15k | TransE | 7.89 | 9.72 | 35.46 |
| | TransH | 7.58 | 10.15 | 37.21 |
| | TransR | 8.55 | 17.09 | 43.80 |
| | TransD | 7.80 | 9.80 | 41.16 |
| WN18 | TransE | 2.52 | 4.37 | 12.25 |
| | TransH | 2.72 | 4.67 | 12.83 |
| | TransR | 2.88 | 3.98 | 13.12 |
| | TransD | 2.94 | 3.88 | 13.03 |

In the future, we will explore other effective measurement to validate a model during training, especially for dense datasets like WN18. We are also interested in clustering of entities and utilizing the graph structure of a knowledge base, which can assist in the generation of negative samples.

ACKNOWLEDGMENT

We thank all anonymous reviewers for their insightful and constructive comments.